\pdfoutput=1

\documentclass[11pt]{article}

\usepackage[preprint]{acl}

\usepackage{times}
\usepackage{latexsym}

\usepackage[T1]{fontenc}

\usepackage[utf8]{inputenc}

\usepackage{microtype}

\usepackage{inconsolata}

\usepackage{microtype}
\usepackage{graphicx}
\usepackage{booktabs}
\usepackage{float}
\usepackage{subfig}

\usepackage{tikz}
\usepackage{pgfplots}
\usepackage{pgfplotstable}

\usepackage{amsmath}
\usepackage{amssymb}
\usepackage{mathtools}
\usepackage{amsthm}
\newcommand{\tensor}[1]{\mathcal{#1}}      
\renewcommand{\vec}[1]{\boldsymbol{#1}}

\usepackage[capitalize,noabbrev]{cleveref}
\theoremstyle{plain}

\theoremstyle{definition}

\theoremstyle{remark}

\newcommand{\minisection}[1]{\noindent{\textbf{#1}}}
\usepackage{multirow}
\usepackage{tabularx}
\usepackage{bbm}
\usepackage{makecell}
\usepackage{enumitem}

\usepackage{moresize}

\title{Parameter-Efficient Sparsity Crafting from Dense to Mixture-of-Experts for Instruction Tuning on General Tasks}

\author{
\textbf{Haoyuan Wu\textsuperscript{$\spadesuit$}},
\textbf{Haisheng Zheng\textsuperscript{$\heartsuit$}},
\textbf{Zhuolun He\textsuperscript{$\spadesuit$,$\clubsuit$}},
\textbf{Bei Yu\textsuperscript{$\spadesuit$}},
\\\\
\textsuperscript{$\spadesuit$}The Chinese University of Hong Kong, Hong Kong SAR \\
\textsuperscript{$\heartsuit$}Shanghai Artificial Intelligent Laboratory, China \\
\textsuperscript{$\clubsuit$}ChatEDA Tech, China\\
\texttt{\{hywu24,byu\}@cse.cuhk.edu.hk} \\
}

\begin{document}
\maketitle

\begin{abstract}
Large language models (LLMs) have demonstrated considerable proficiency in general natural language processing (NLP) tasks. 
Instruction tuning, a successful paradigm, enhances the ability of LLMs to follow natural language instructions and exhibit robust generalization across general tasks. 
However, these models often encounter performance limitations across multiple tasks due to constrained model capacity. 
Expanding this capacity during the instruction tuning phase poses significant challenges. 
To address this issue, we introduce parameter-efficient sparsity crafting (PESC), which crafts dense models into sparse models using the mixture-of-experts (MoE) architecture. 
PESC integrates adapters into the MoE layers of sparse models, differentiating experts without altering the individual weights within these layers. 
This method significantly reduces computational costs and GPU memory requirements, facilitating model capacity expansion through a minimal parameter increase when guaranteeing the quality of approximation in function space compared to original sparse upcycling. 
Our empirical evaluation demonstrates the effectiveness of the PESC method. 
Using PESC during instruction tuning, our best sparse model outperforms other sparse and dense models and exhibits superior general capabilities compared to GPT-3.5.
Our code is available at \url{https://github.com/wuhy68/Parameter-Efficient-MoE}.
\end{abstract}

\begin{figure}[tb!]
    \centering
    \includegraphics[width=0.888\linewidth]{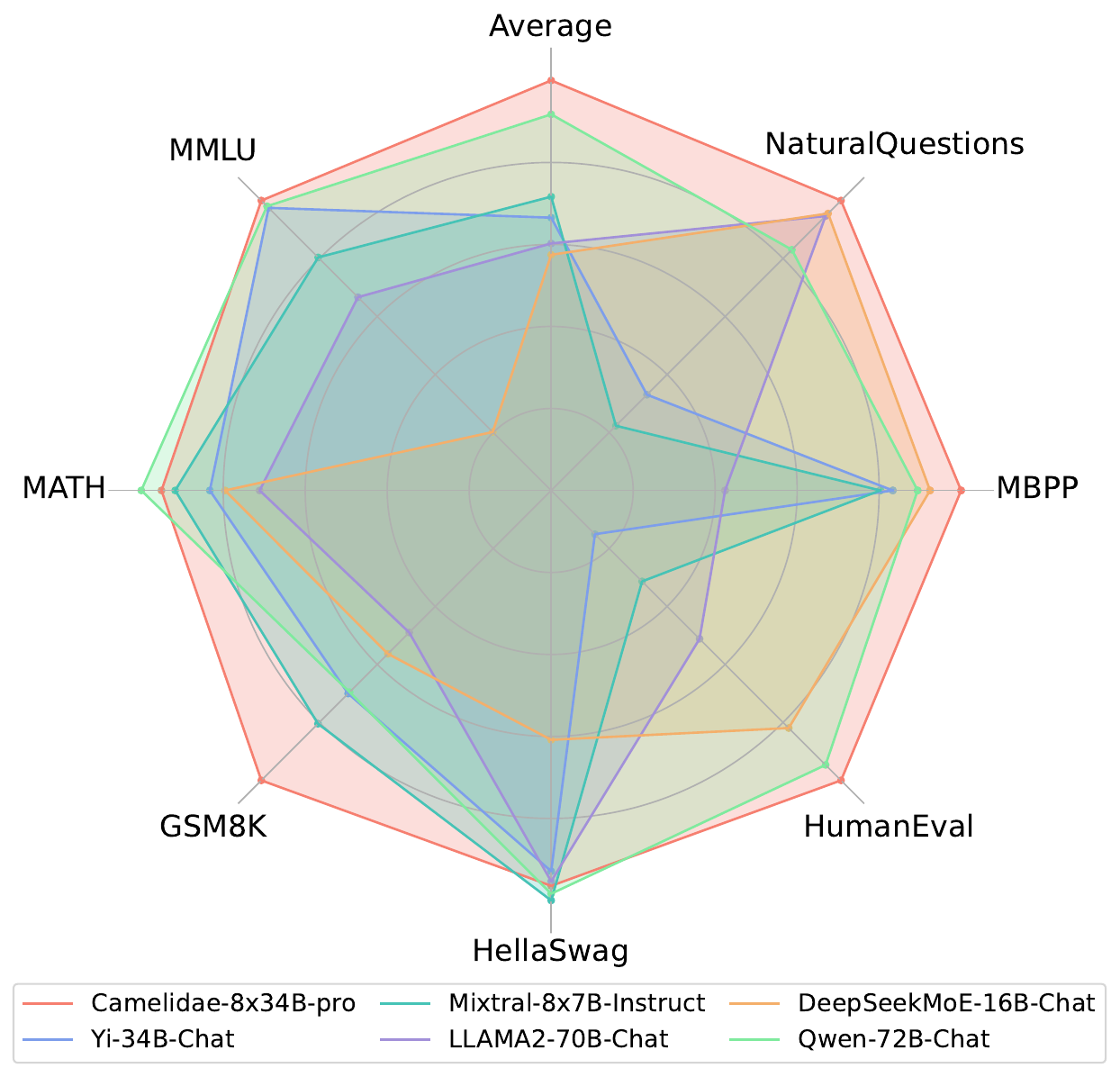} 
    \caption{Camelidae-8$\times$34B-pro achieves excellent performance across general tasks.}
    \label{fig:radar}
\end{figure}

\section{Introduction}

Recent advancements in NLP have been significantly propelled by the advent of LLMs such as GPT \cite{brown2020language, openai2023gpt4}, Llama \cite{touvron2023llama, touvron2023llama2}, Mistral \cite{2023mistral, jiang2024mixtral}, etc.
The increasing scale of LLMs has established them as the experts for NLP tasks due to their exceptional ability to identify complex linguistic patterns \cite{wei2022emergent}.

A prominent method for training LLMs is instruction tuning \cite{wei2021finetuned}. This approach utilizes large-scale, well-formatted instruction data, enabling LLMs to refine their pre-trained representations to comply with human instructions \cite{Rohan2023alpaca, xu2023wizardlm, dettmers2024qlora, mukherjee2023orca}. 
Such instruction-tuned LLMs exhibit remarkable generalization capabilities in NLP tasks \cite{longpre2023flan}. 
This generalization requires training on a broad range of instruction-following tasks from multiple domains such as math, code, biology, etc \cite{chung2022scaling, sanh2021multitask}. 
However, the inherent complexity of these tasks can hinder model fine-tuning \cite{zhang2021survey}.
Specifically, models of certain sizes may struggle to optimize losses from conflicting tasks, resulting in subpar performance for general tasks.

The scaling law \cite{chung2022scaling} suggests that increasing the model's scale is crucial for better performance.
Expanding the model's capacity can also improve instruction tuning effectiveness for general tasks \cite{kaplan2020scaling}. 
Nonetheless, most LLMs are pre-trained dense models designed based on transformer architecture, which limits scalability during instruction tuning. 
\citet{komatsuzaki2022sparse} presented a method for upcycling dense models into sparse activated MoE models, which boast greater capacity \cite{shazeer2017outrageously, lepikhin2020gshard, fedus2022switch, puigcerver2023sparse}. 
Notably, \citet{shen2023mixture} suggested that MoE models respond more effectively to instruction tuning compared to dense models. 
Consequently, converting dense models into MoE models during instruction tuning has the potential to achieve great performance on general tasks. 
This conversion involves initializing each expert in the MoE models as a copy of the feedforward neural network (FFN) layers \cite{chen2015net2net, rae2021scaling}. 
Given the parameter scale of current LLMs, training such giant models requires updating the weights of experts in the MoE layer, which is constrained by GPU memory resources and computational costs. 

To mitigate these challenges, we introduce parameter-efficient sparsity crafting (PESC), an approach that effectively expands model capacity while synergizing with parameter-efficient fine-tuning (PEFT) techniques \cite{houlsby2019parameter, dettmers2024qlora}.
PESC involves inserting adapters \cite{houlsby2019parameter} into the MoE layers of sparse models, allowing differentiation between experts without altering each expert's weights in the MoE layers when guaranteeing the quality of the approximation in function space compared to original sparse upcycling \cite{komatsuzaki2022sparse}. 
Considering that the more sophisticated construction can improve the approximation \cite{ding2022delta}, we also apply the QLoRA \cite{dettmers2024qlora} technique to update other weights in the sparse models.
As shown in \Cref{fig:radar}, our Camelidae-8$\times$34B-pro, instruction fine-tuned utilizing PESC, achieved the best performance among various open-source sparse models and dense models.
Our contributions are described as follows:
\begin{itemize}[itemsep=0pt,topsep=0pt,parsep=0pt]
    \item We propose an approach, parameter-efficient sparsity crafting (PESC), for the extension of the model capacity efficiently.
    \item We implement the PESC method for instruction tuning across general tasks, achieving significant performance improvements on various benchmarks.
    \item We develop Camelidae models, sparse models trained with the PESC method, achieving the best performance across open-source sparse models and demonstrating superior general capabilities compared to GPT-3.5.
\end{itemize}

\section{Methodology}

\subsection{Preliminaries}

\minisection{Adapters.}
\citet{houlsby2019parameter} proposed the integration of adapters into pre-trained transformer-based models to enhance parameter efficiency. 
This approach involves tuning only the parameters added by the adapters. 
An adapter consists of two matrices, $\vec{W}_{\text{down}} \in \mathbb{R}^{d_1 \times d_2}$ and $\vec{W}_{\text{up}} \in \mathbb{R}^{d_2 \times d_1}$, coupled with a non-linear function $\sigma(\cdot)$. 
Here, $d_1$ and $d_2$ denote the feature dimensions in the pre-trained models and the adapter's hidden dimension, respectively, with $d_2 < d_1$ typically. 
Given a feature $\vec{U} \in \mathbb{R}^{N \times d_1}$ in the pre-trained model, the output of the Adapter module is expressed as:
\begin{equation}
    \vec{U}' = \sigma(\vec{U}\vec{W}_{\text{down}})\vec{W}_{\text{up}} + \vec{U}.
\label{eq:adapter}
\end{equation}

\begin{figure}[tb!]
    \centering
    \includegraphics[width=\linewidth]{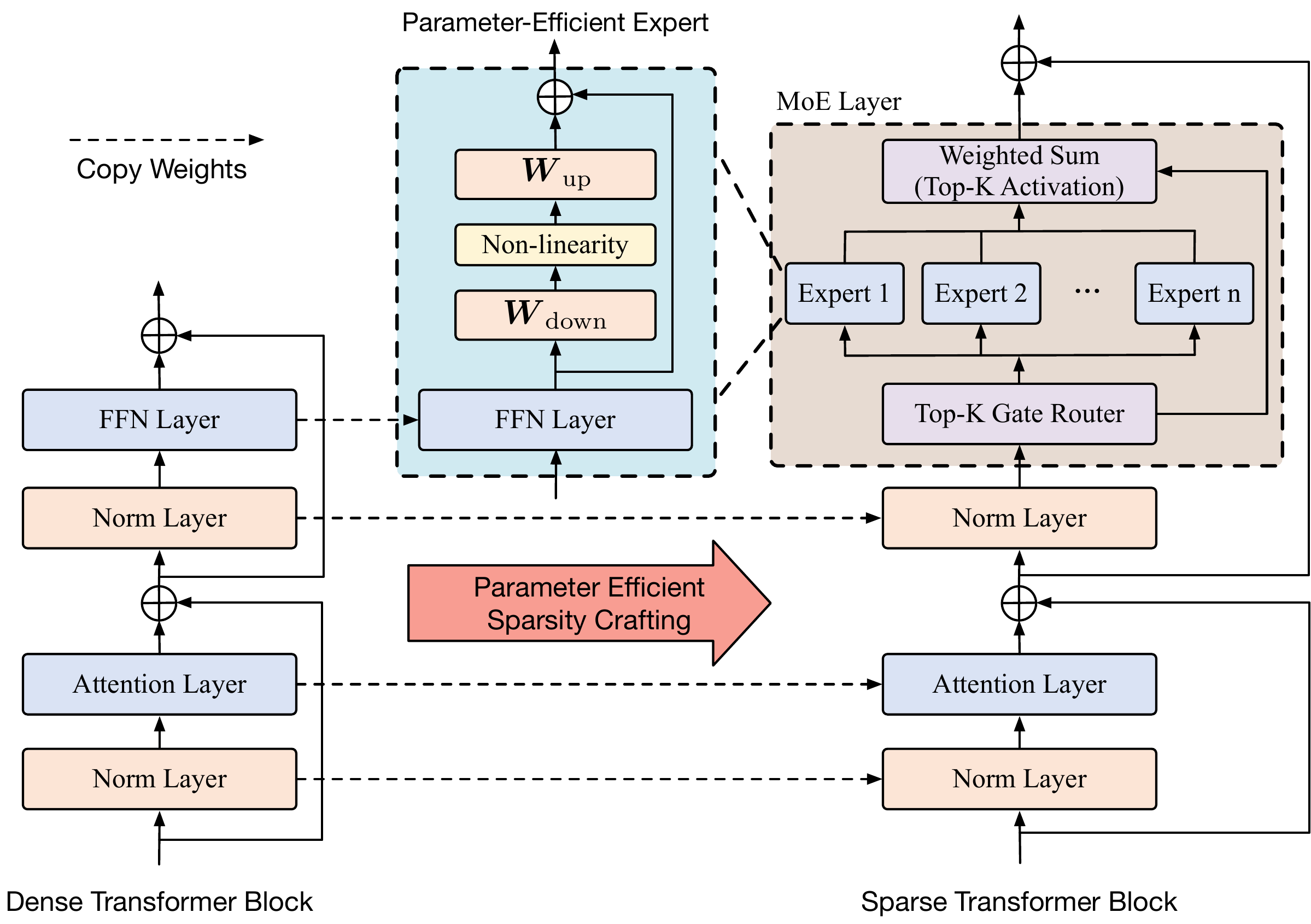} 
    \caption{Overview of the parameter-efficient sparsity crafting with parameter-efficient experts.}
    \label{fig:sparsity_crafting}
\end{figure}

\minisection{Mixture-of-Experts.}
As depicted in \Cref{fig:sparsity_crafting}, an MoE layer comprises $n$ experts, $\{E_i\}_{i=1}^n$, and a router $R$. 
The output $\vec{y}$ for an input $\vec{x}$ in the MoE layer is computed as:
\begin{equation}
    \vec{y} = \sum_{i=1}^n R(\vec{x})_iE_i(\vec{x}),
\label{eq:moe}
\end{equation}
where $R(\vec{x})_i$ represents the output of the gating network for the $i$-th expert, and $E_i(\vec{x})$ is the output of the $i$-th expert.

\minisection{Sparsity Crafting.}
Building on the concept of sparsity upcycling \cite{komatsuzaki2022sparse}, sparsity crafting leverages the weights of dense models. 
As depicted in \Cref{fig:sparsity_crafting}, sparsity crafting involves a transformative process: substituting the FFN layer $F$ within each block of the dense transformer model with an MoE layer. 
This replacement gives rise to an innovatively sparse transformer block. 
During the initialization phase of sparsity crafting, each expert $E_i$ within the MoE layer is initialized with the FFN layer $F$. 
To ensure structural coherence, other components, such as the normalization and attention layers, are replicated directly from the dense transformer block.

For clarity, let us define $\mathcal{F}_{i}(\theta_{i})$ as the objective function for the $i$-th expert in the MoE layer, where $\theta_{i}$ represents the parameters for $E_{i}$. 
$\tensor{\theta}_{i}$ is initialized from $\theta_{o}$, which are the parameters of the FFN layer $F$ from the original dense model. 
The essence of the sparsity crafting training regimen lies in the optimization of $\mathcal{F}_{i}(\theta_{i})$. 
The goal is to derive $\theta_{i}^{+}$, the optimized parameters for each expert. 
This is formally expressed as:
\begin{equation}
    \theta_{i}^{+} = \arg \min_{\theta_{i}} \mathcal{F}_{i}(\tensor\theta_{i}).
\label{eq:sc}
\end{equation}
After the instruction tuning process utilizing the sparsity crafting technique, the optimized parameter sets $\{\theta_{i}^{+}\}_{i=1}^{n}$ are obtained for experts $\{E_{i}\}_{i=1}^{n}$ in the MoE layer.

\begin{figure}[tb!]
    \centering
    \includegraphics[width=0.748\linewidth]{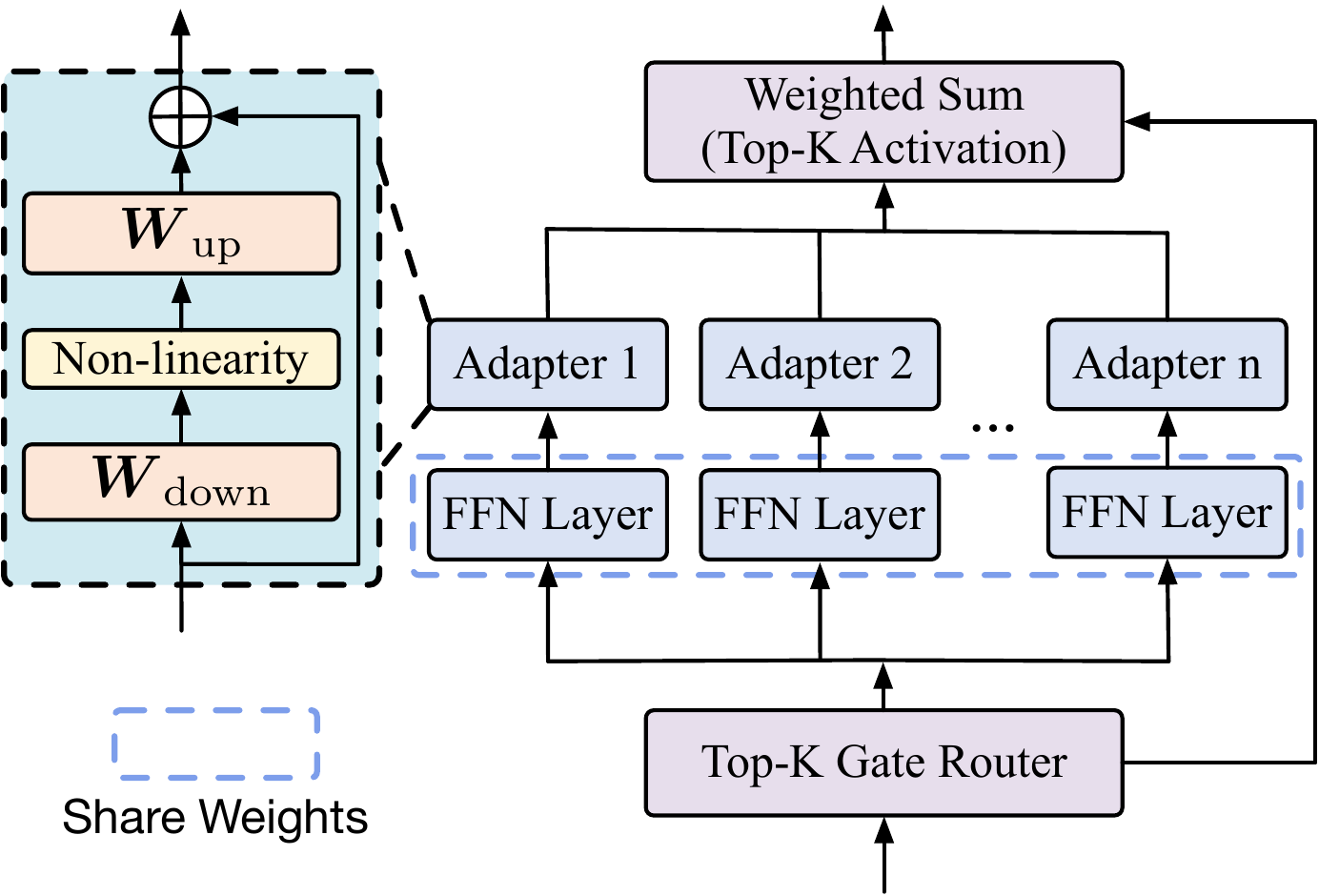} 
    \caption{Detailed design of the MoE layer for PESC utilizing parameter-efficient experts.
    All the FFN layers share the same weights.}
    \label{fig:moe_layer}
\end{figure}

\subsection{Parameter-Efficient Sparsity Crafting}
\label{sec:pesc}
As shown in \Cref{eq:sc}, traditional sparsity crafting necessitates optimizing the parameters $\{\theta_{i}\}_{i=1}^{n}$ for each expert $E_{i}$ in the MoE layer, leading to significant resource consumption, including training time and memory costs due to the extensive parameters of FFN layers in LLMs. 
Consequently, as illustrated in \cref{fig:sparsity_crafting}, we introduce PESC, an approach that addresses the high training time and memory costs associated with sparsity crafting in LLMs.
Specifically, PESC, leveraging the parameter-efficient fine-tuning (PEFT) paradigm, focuses on tuning a smaller subset of parameters to achieve efficiency.

The core of PESC lies in its objective function, $\tilde{\mathcal{F}_{i}}(\theta_{i}, \omega_{i})$, where $\omega_{i}$ represents the select parameters for tuning. 
Notably, the parameters of $\omega_{i}$ is significantly less than $\theta_{i}$, as indicated by $|\omega_{i}| \ll |\theta_{i}|$, where $|\cdot|$ indicates the number of parameters involved.
Each expert $E_{i}$ begins the process with the initial state $(\theta_{o}, \omega_{o})$, where $\omega_{o}$ is initialized to zero to facilitate identity mapping, resulting in $\tilde{\mathcal{F}_{i}}(\theta_{o}, \omega_{o}) = \mathcal{F}_{i}(\theta_{o})$. 
The training procedure for PESC is thus the optimization of $\tilde{\mathcal{F}}_{i}(\theta_{o}, \omega_{i})$, leading to a solution $\omega_{i}^{+}$ defined as:
\begin{equation}
    \omega_{i}^{+} = \arg \min_{\omega_{i}} \tilde{\mathcal{F}}_{i}(\theta_{o}, \omega_{i}).
\label{eq:pesc}
\end{equation}
Considering that $|\omega_{i}| \ll |\theta_{i}|$, we have
\begin{align}
    \sum_{i=1}^n |\omega_{i}^{+}| + |\theta_{o}| &= n \times |\omega_{o}| + |\theta_{o}| \nonumber\\
    &\ll n \times |\theta_{o}| = \sum_{i=1}^n |\theta_{i}^{+}|.
\label{eq:size}
\end{align}
Consequently, this solution set $\{\omega_{i}^{+}\}_{i=1}^{n}$ is more efficient than the original sparsity crafting parameters $\{\theta_{i}^{+}\}_{i=1}^{n}$ for the set $\{E_{i}\}_{i=1}^{n}$.

To ensure the effectiveness of PESC compared to traditional sparsity crafting, it is vital to maintain a small approximation error, as defined by:
\begin{equation}
    \lvert\tilde{\mathcal{F}}_{i}(\theta_{i}^{+}, \omega_{o}) - \tilde{\mathcal{F}}_{i}(\theta_{o}, \omega_{i}^{+})\rvert < \xi,
\label{eq:err}
\end{equation}
where $\xi$ is the approximation error.
This can be achieved by designing an approximate function $\tilde{\mathcal{F}}_{i}(\theta_{o}, \omega_{i}^{+})$ that closely matches $\tilde{\mathcal{F}}_{i}(\theta_{i}^{+}, \omega_{o})$ \cite{houlsby2019parameter, ding2022delta}. 
Considering that the trajectory of $\theta_{i}$ optimization approximately follows a manifold, which can be projected into a lower-dimensional space such as adapter in \Cref{eq:adapter}.
The approximation error is contingent on the representational capacity of the inserted adapters. 
Given the universal approximation property of MLP layers with general activation functions, the Adapter module is a universal approximator \cite{funahashi1989approximate, leshno1993multilayer, kidger2020universal}.
As a result, utilizing the adapters as $\omega_{i}$ can effectively ensure the quality of the approximation of $\tilde{\mathcal{F}}_{i}(\theta_{i}^{+}, \omega_{o})$.

\subsection{Model Design}

\minisection{Parameter-Efficient Experts.}
According to the analysis in \Cref{sec:pesc}, adapters can guarantee a good lower bound $\xi$ in \Cref{eq:err}.
Consequently, we can introduce parameter-efficient MoE layers by integrating adapters, thereby achieving sparsity in a more parameter-efficient manner.

In the training of sparse transformer blocks, gradients are back-propagated to each expert, necessitating parameter updates.
For a collection of $n$ experts, original sparsity crafting demands a computational cost $n$ times that of a single FFN layer.
As depicted in \Cref{fig:moe_layer}, our PESC utilizes adapters to circumvent redundant updates of the expert weights $\theta_{i}$.
Specifically, we update the $\omega_{i}$ of $n$ inserted adapters to differentiate between experts without altering each expert's original weights $\theta_{o}$ replicated from the original FFN layer. 
Thus, for a given input $\vec{x}$, \Cref{eq:moe} can be reformulated as:
\begin{equation}
    \vec{y} = \sum_{i=0}^{n} R(\vec{x})_{i}A_{i}(E(\vec{x})),
\label{eq:moe_re}
\end{equation}
where $A_{i}(\vec{x})$ construct the parameter-efficient expert as follows:
\begin{equation}
    A_{i}(\vec{x}) = \sigma(\vec{x}{\vec{W}_{i}}_{\text{down}}){\vec{W}_{i}}_{\text{up}} + \vec{x}.
\end{equation}
Considering that the more sophisticated construction can improve the approximation, we can also update the shared weights $\theta_{o}$ of $\{E_i\}_{i=1}^n$.
As illustrated in \Cref{eq:moe_re}, this approach allows for efficient scaling of the model capacity by introducing a minimal number of parameters across $n$ inserted adapters.

\minisection{Top-K Gate Router.}
Within the sparse transformer block, the MoE layer encompasses a specified number of experts.
A router, employing a softmax activation function, models a probability distribution over these experts, reflecting each expert's capability to process incoming tokens.
The router's weights, denoted as $\vec{W}_{r}$, which are integrated into the sparse transformer block, are initially randomly initialized. 
As depicted in \Cref{fig:moe_layer}, we utilize the top-k gate router within the sparse transformer block \cite{lepikhin2020gshard, du2022glam}.
This router activates the most suitable two experts out of $n$ experts $\{E_{i}\}_{i=1}^{n}$ for each token $\vec{x}$ in an input sequence.
After receiving the input token $\vec{x}$, the router produces router logits $R(\vec{x}) = \vec{W}_{r} \cdot \vec{x}$.
Before being normalized via a softmax distribution over the available $n$ experts, we perform the KeepTopK function.
The KeepTopK function is applied to retain only the top-k values of the router logits, assigning $-\infty$ to the rest, effectively zeroing them post-softmax normalization.
Thus, given a token $\vec{x}$, the router's output logit is represented as:
\begin{equation}
    R(\vec{x}) = \text{Softmax}(\text{KeepTopK}(\vec{W}_{r} \cdot \vec{x})).
\end{equation}
The gate value of each expert $E_{i}$ for the input token $\vec{x}$ is $R(\vec{x})_i$.
Despite an increase in parameters, the experts of the MoE layer are activated sparsely, implying that only a limited subset of experts is used per input token. 
This approach enhances the capacity of the model while maintaining computational efficiency.
The top-k gate router selects the best two experts for each token during inference. 
In an MoE layer with $n$ experts, this enables up to $\binom{n}{k}$ different combinations of experts, as opposed to a single combination in the traditional transformer architecture, providing enhanced computational adaptability. 

\minisection{Experts Loading Balance.}
The top-k gate router, through its gating mechanism, tends to disproportionately favor a few experts, leading to an imbalance where these experts are more frequently trained and consequently chosen by the router. 
To counter this imbalance and promote uniform expert utilization, an auxiliary loss as suggested by \citet{fedus2022switch} is integrated during training for each sparse transformer block.
With $n$ experts and a batch $B$ containing $T$ tokens, this auxiliary loss $\mathcal{L}$ for experts loading balance is calculated as the scaled dot-product of vectors $\vec{f}$ and $\vec{p}$,
\begin{equation}
\mathcal{L} = \alpha \cdot n \cdot \sum_{i=1}^{n} {\Vec{f}}_i \cdot \Vec{p}_i,
\label{eq:balance_loss}
\end{equation}
where ${f}_i$ denotes the fraction of tokens dispatched to expert $i$ and ${p}_i$ represents the fraction of router probability allocated to expert $i$. 
$\alpha$ is a multiplicative coefficient for the auxiliary losses.
We utilize an $\alpha=10^{-2}$ which was sufficiently large to ensure load balancing while small enough to not overwhelm the primary cross-entropy objective.
As the ideal scenario entails uniform routing across the $n$ experts, both vectors should ideally have values of $\frac{1}{n}$. 
The auxiliary loss of \Cref{eq:balance_loss} fosters this uniform distribution, achieving its minimum under such conditions.

\section{Experiments}
\label{sec:exp}

\subsection{Settings}

\minisection{Training Data.}
To demonstrate the learning ability of the sparse model with MoE layers, we simultaneously trained the model on a diverse set of skills, encompassing coding, mathematical, and other general abilities from various subjects.
This training involved integrating three distinct datasets from varied domains during the instruction tuning phase: SlimOrca \cite{SlimOrca, mukherjee2023orca, longpre2023flan}, Magicoder \cite{wei2023magicoder}, and MetaMathQA \cite{yu2023metamath} datasets. 
After filtration and sampling, we can get two instruction datasets including IDAE-500K and IDAE-720K finally.
We provide more details of IDAE datasets in \Cref{sec:appendix1}.

\begin{table*}[t]
\centering
\label{table:main_1}
\renewcommand{\arraystretch}{1.0}
\resizebox{0.868\linewidth}{!}{
\scriptsize
\begin{tabularx}{0.838\linewidth}{X|ccc|cccc}
\toprule
& \multicolumn{3}{c|}{Sparse Chat Models} & \multicolumn{4}{c}{Dense Chat Models} \\
\cmidrule{2-4} \cmidrule{5-8}
& \makecell{Camelidae \\ 8$\times$34B-pro} & \makecell{Mixtral \\ 8$\times$7B Inst.} & 
\makecell{DeepSeekMoE \\ 16B Chat}  & \makecell{Yi \\ 34B Chat} & \makecell{Llama2 \\ 70B Chat} & \makecell{Qwen \\ 72B Chat} & GPT-3.5 \\
\midrule
\makecell[X]{MMLU (Acc.) \newline \cite{hendrycks2020measuring}}
    & \makecell{\textbf{75.7\%} \\ (5-shot)} & \makecell{68.7\% \\ (5-shot)} & \makecell{47.2\% \\ (5-shot)} & \makecell{74.8\% \\ (5-shot)} & \makecell{63.8\% \\ (5-shot)} & \makecell{75.0\% \\ (5-shot)} & \makecell{70.0\% \\ (5-shot)} \\ 
\midrule
\makecell[X]{GSM8K (Acc.) \newline \cite{cobbe2021training}}      
    & \makecell{\textbf{79.4\%} \\ (5-shot)} & \makecell{71.7\% \\ (5-shot)} & \makecell{62.2\% \\ (5-shot)} & \makecell{67.6\% \\ (5-shot)} & \makecell{59.3\% \\ (5-shot)} & \makecell{67.4\% \\ (5-shot) } & \makecell{57.1\% \\ (5-shot)} \\ 
\midrule
\makecell[X]{MATH (Acc.) \newline \cite{hendrycks2021measuring}}      
    & \makecell{24.0\% \\ (4-shot)} & \makecell{22.1\% \\ (4-shot)} & \makecell{15.2\% \\ (4-shot)} & \makecell{17.3\% \\ (4-shot)} & \makecell{10.4\% \\ (4-shot)} & \makecell{26.8\% \\ (4-shot)} & \makecell{\textbf{34.1}\% \\ (4-shot)}\\ 
\midrule
\makecell[X]{HumanEval (Pass@1) \newline \cite{chen2021evaluating}}   
    & \makecell{\textbf{48.8\%} \\ (0-shot)} & \makecell{25.6\% \\ (0-shot)} & \makecell{42.7\% \\ (0-shot)} & \makecell{20.1\% \\ (0-shot)} & \makecell{32.3\% \\ (0-shot)} & \makecell{47.0\% \\ (0-shot)} & \makecell{48.1\% \\ (0-shot)} \\ 
\midrule
\makecell[X]{MBPP (Pass@1) \newline (\cite{austin2021program}}      
    & \makecell{\textbf{43.2\%} \\ (4-shot)} & \makecell{40.6\% \\ (4-shot)} & \makecell{42.2\% \\ (4-shot)} & \makecell{41.0\% \\ (4-shot)} & \makecell{35.6\% \\ (4-shot)} & \makecell{41.8\% \\ (4-shot)} & - \\ 
\midrule
\makecell[X]{HellaSwag (Acc.) \newline \cite{zellers2019hellaswag}}
    & \makecell{85.2\% \\ (10-shot)} & \makecell{\textbf{86.5}\% \\ (10-shot)} & \makecell{72.2\% \\ (10-shot)} & \makecell{83.9\% \\ (10-shot)} & \makecell{84.8\% \\ (10-shot)} & \makecell{85.9\% \\ (10-shot)} & \makecell{85.5\% \\ (10-shot)} \\
\midrule
\makecell[X]{NaturalQuestions (EM) \newline \cite{kwiatkowski2019natural}}   
    & \makecell{\textbf{31.2\%} \\ (0-shot)} & \makecell{22.5\% \\ (0-shot)} & \makecell{30.7\% \\ (0-shot)} & \makecell{23.7\% \\ (0-shot)} & \makecell{30.6\% \\ (0-shot)} & \makecell{29.3\% \\ (0-shot)} & - \\
\bottomrule
\end{tabularx}}
\caption{Performance of Camelidae-8$\times$34B-pro on academic benchmarks. 
We present a detailed comparison of the Camelidae-8$\times$34B-pro model with the various open-source sparse chat models and dense chat models. 
We bold the highest scores among all models.}
\end{table*}

\minisection{Evaluation Benchmarks.}
Our evaluation compares the performance of dense and sparse models on academic benchmarks. 
The dense models include Llama2 \cite{touvron2023llama2}, Vicuna \cite{zheng2023judging}, Yi \cite{2023Yi}, SUSChat \cite{2023SUSChat}, Qwen \cite{bai2023qwen}, GPT3.5 \cite{brown2020language}, and our Camel models, while the sparse models encompass Mixtral \cite{jiang2024mixtral}, DeepSeekMoE \cite{dai2024deepseekmoe}, and our Camelidae models. 
Evaluations are conducted using OpenCompass \cite{2023opencompass}, LM-Eval-Harness \cite{eval-harness}, and our internal evaluation libraries, summarizing performances across well-known benchmarks. 
These benchmarks are illustrated as follows:
\begin{itemize}[itemsep=0pt,topsep=0pt,parsep=0pt]
    \item \textbf{Code:} Evaluation includes pass@1 scores for HumanEval \cite{chen2021evaluating} and MBPP \cite{austin2021program}.
    \item \textbf{Math:} Accuracy scores for GSM8K \cite{cobbe2021training} (5-shot) and MATH \cite{hendrycks2021measuring} (4-shot) benchmarks.
    \item \textbf{Commonsense Reasoning (CR):} Accuracy scores for PIQA \cite{bisk2020piqa}, HellaSwag \cite{zellers2019hellaswag}, WinoGrande \cite{sakaguchi2021winogrande}, ARC-easy, and ARC-challenge \cite{clark2018think}.
    \item \textbf{Word Knowledge (WK):} Assessment of 0-shot performance on NaturalQuestions \cite{kwiatkowski2019natural} and TriviaQA \cite{joshi2017triviaqa} utilizing the exact match (EM) metric.
    \item \textbf{Aggregated Benchmarks:} Overall results for MMLU \cite{hendrycks2020measuring} (5-shot) utilizing accuracy scores metrics.
\end{itemize}
Notably, for more detailed experiment results, please refer to \Cref{sec:appendix3}.

\minisection{Camel and Camelidae Models.}
We fine-tuned Camel and Camelidae models using identical datasets, IDAE-500K, to ensure fair comparisons between dense and sparse models.
Specifically, Camel models are dense models while Camelidae models are sparse models with MoE architecture.
Notably, to further enhance the capabilities of the sparse models, we also utilize IDAE-720K for the instruction-tuning of the Camelidae-pro model.
All Camelidae models utilize the top-2 gate router. 

\minisection{Implementation Details.}
We employed QLoRA \cite{dettmers2024qlora} techniques for effective fine-tuning of both the Camel and Camelidae models derived from Llama2-7B \cite{touvron2023llama2}, Llama2-13B \cite{touvron2023llama2}, and Yi-34B \cite{2023Yi}. 
As for the QLoRA configuration, we used a 4-bit quantization scheme for our experiments, which significantly reduces memory usage while preserving model performance.
This process entailed using a constant learning rate schedule with a warm-up ratio of 0.03, and the paged AdamW \cite{dettmers2024qlora, loshchilov2017decoupled} optimizer with a learning rate of \(2 \times 10^{-4}\), no weight decay, a batch size of 128, and a sequence length of 2048 tokens. 
The models underwent instruction tuning for one epoch on 16 A100 GPUs, each equipped with 80G memory.
Please refer to \Cref{sec:appendix2} for more details.

\subsection{Comparison with Chat LLMs}

\begin{table*}[t]
\centering
\renewcommand{\arraystretch}{1.0}
\resizebox{0.868\linewidth}{!}{
\scriptsize
\begin{tabularx}{0.838\linewidth}{l|cc|cc|ccc}
\toprule
& \makecell{Camel-7B} & \makecell{Camelidae \\ 8$\times$7B} & \makecell{Camel-13B} & \makecell{Camelidae \\ 8$\times$13B} & \makecell{Camel-34B} & \makecell{Camelidae \\ 8$\times$34B} & \makecell{Camelidae \\ 8$\times$34B-pro} \\ 
\midrule
\# Total Params & 7B & 8B & 13B & 15B & 34B & 38B & 38B \\
\# Activated Params & 7B & 7B & 13B & 14B & 34B & 35B & 35B \\
\# Training Instructions & 500K & 500K & 500K & 500K & 500K & 500K & 720K \\
\midrule
MMLU (Acc.) & 47.7 & \textbf{48.3} & \textbf{54.4} & \textbf{54.4} & 75.3 & 75.6 & \textbf{75.7} \\
\midrule
HumanEval (Pass@1) & 17.7 & \textbf{18.3} & 28.7 & \textbf{30.6} & 42.1 & 43.9 & \textbf{48.8} \\
MBPP (Pass@1) & 21.0 & \textbf{23.4} & 30.3 & \textbf{30.4} & 40.6 & 41.4 & \textbf{43.2} \\
\midrule
GSM8K (Acc.) & 40.7 & \textbf{44.0} & 50.2 & \textbf{52.6} & 76.1 & 78.3 & \textbf{79.4} \\
MATH (Acc.) & 4.8 & \textbf{5.8} & 8.4 & \textbf{9.8} & 18.2 & 22.6 & \textbf{24.0} \\
\midrule
PIQA (Acc.) & 79.7 & \textbf{79.9} & \textbf{80.9} & \textbf{80.9} & 82.3 & 82.7 & \textbf{83.6} \\
HellaSwag (Acc.) & \textbf{76.8} & \textbf{76.8} & 79.8 & \textbf{80.1} & 82.6 & \textbf{83.2} & 82.5 \\
Winogrande (Acc.) & 71.3 & \textbf{72.1} & 74.6 & \textbf{74.7} & 80.0 & \textbf{80.9} & 80.1 \\
ARC-easy (Acc.) & \textbf{75.0} & \textbf{75.0} & 77.7 & \textbf{78.8} & 86.1 & 86.2 & \textbf{86.6} \\
ARC-challenge (Acc.) & 47.9 & \textbf{49.6} & \textbf{54.3} & 54.2 & 63.6 & \textbf{65.2} & 63.3 \\
\midrule
NaturalQuestions (EM) & 17.6 & \textbf{17.8} & 24.7 & \textbf{26.8} & 31.6 & \textbf{32.2} & 31.2 \\
TriviaQA (EM) & \textbf{51.0} & \textbf{51.0} & 57.5 & \textbf{59.4} & 63.3 & \textbf{63.4} & 62.5 \\
\bottomrule
\end{tabularx}}
\caption{Overall performance on all the evaluation benchmarks of dense models (Camel) and sparse (Camelidae) models across different model sizes. 
We bold the highest scores separately for different model sizes.}
\label{table:main_2}
\end{table*}

We present the performance of various chat LLMs on a set of standardized benchmarks. 
The chat models evaluated are Camelidae-8$\times$34B-pro, Mixtral-8$\times$7B-Instruct \cite{jiang2024mixtral}, DeepSeekMoE-16B-Chat \cite{dai2024deepseekmoe}, Yi-34B-Chat \cite{2023Yi}, Llama2-70B-Chat \cite{touvron2023llama2}, Qwen-72B-Chat \cite{bai2023qwen}, and GPT-3.5 \cite{brown2020language}. 
The benchmarks cover a range of domains, including multiple-choice questions across 57 subjects (MMLU), grade-school math (GSM8K), math problems across various difficulty levels (MATH), Python coding tasks (HumanEval), Python code generation (MBPP), commonsense reasoning (HellaSwag), and world knowledge question answering (NaturalQuestions).

As shown in \Cref{table:main_1}, Camelidae-8$\times$34B-pro demonstrates its strengths in its wide range of knowledge, mathematical, coding, and commonsense reasoning capabilities across various sparse and dense models.

\minisection{Knowledge and Reasoning Abilities.} 
Camelidae-8$\times$34B-pro demonstrates impressive performance on MMLU with a high success rate of 75.7\%, indicating its wide-ranging professional and academic knowledge.
Meanwhile, Camelidae-8$\times$34B-pro scores 31.2\% on NaturalQuestions, demonstrating a comprehensive world knowledge base.
Although Camelidae-8$\times$34B-pro is weaker than some models in the HellaSwag benchmark, its 85.2\% accuracy is still decent for commonsense reasoning.

\minisection{Mathematical Proficiency.} 
Camelidae-8$\times$34B-pro excels on the GSM8K benchmark with 79.4\% accuracy, the highest among models. 
However, its 24.0\% score on the MATH benchmark lags behind GPT-3.5, indicating a relative weakness in solving more complex mathematical problems.

\minisection{Coding Skills.} 
Camelidae-8$\times$34B-pro demonstrates strong coding abilities with 48.8\% accuracy on the HumanEval benchmark, comparable to GPT-3.5, and a 43.2\% pass rate on the MBPP Python code generation benchmark, showcasing its prowess in understanding and generating code.

\subsection{Ablation Studies}

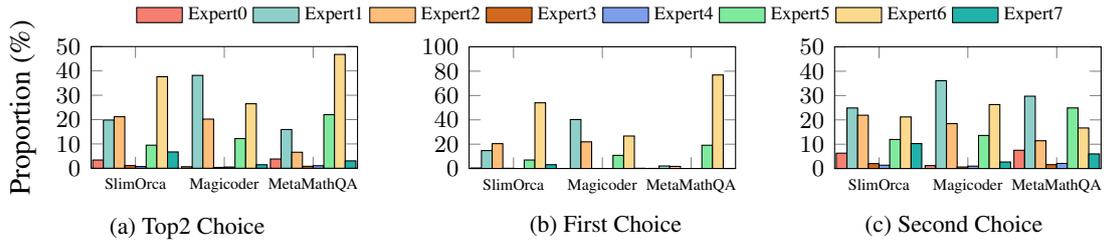
\begin{figure*}
    \centering
    \begin{minipage}[]{0.318\linewidth}
        \vspace{0.215cm}
        \hspace{0.1cm}
        \subfloat[Top2 Choice]{\pgfplotsset{
    width = \textwidth,
    height = 0.628\textwidth
}

\begin{tikzpicture}[scale=1.0]

    \definecolor{expert0}{HTML}{F67F70}
    \definecolor{expert1}{HTML}{8ECFC9}
    \definecolor{expert2}{HTML}{FFBE7A}
    \definecolor{expert3}{HTML}{D2691E}
    \definecolor{expert4}{HTML}{7D9EEB}
    \definecolor{expert5}{HTML}{7FEA9E}
    \definecolor{expert6}{HTML}{FBD98D}
    \definecolor{expert7}{HTML}{20B2AA}
    
    \begin{axis}[
        ybar=0pt,
        xmin=0.5, xmax=3.5,
        ymin=0, ymax=50,
        xticklabels={SlimOrca, Magicoder, MetaMathQA},
        xtick={1,2,3},
        xtick align=inside,
        xticklabel style = {rotate=0}, 
        xticklabel style = {font=\tiny},
        ytick={0,10,20,30,40,50},
        yticklabel style = {font=\footnotesize},
        bar width = 4pt,
        ylabel={Proportion (\%)},
        ylabel near ticks,
        legend style={
            draw=none,
            at={(0.5,1.1)},
            anchor=south,
            legend columns=4,
        }
        ]
        \addplot+[ybar, fill=expert0, draw=black, area legend]
        table [x={Expert},  y={Data}]     {data/Top2_Expert0.dat};
        \addplot+[fill=expert1, draw=black, area legend]         
        table [x={Expert},  y={Data}]     {data/Top2_Expert1.dat};
        \addplot+[fill=expert2, draw=black, area legend]         
        table [x={Expert},  y={Data}]     {data/Top2_Expert2.dat};
        \addplot+[fill=expert3, draw=black, area legend]         
        table [x={Expert},  y={Data}]     {data/Top2_Expert3.dat};
        \addplot+[fill=expert4, draw=black, area legend]         
        table [x={Expert},  y={Data}]     {data/Top2_Expert4.dat};
        \addplot+[fill=expert5, draw=black, area legend]         
        table [x={Expert},  y={Data}]     {data/Top2_Expert5.dat};
        \addplot+[fill=expert6, draw=black, area legend]         
        table [x={Expert},  y={Data}]     {data/Top2_Expert6.dat};
        \addplot+[fill=expert7, draw=black, area legend]   
        table [x={Expert},  y={Data}]     {data/Top2_Expert7.dat};
    \end{axis}
\end{tikzpicture}
        \label{fig:top2_choice}}
    \end{minipage}
    \begin{minipage}[]{0.318\linewidth}
        \vspace{0.cm}
        \hspace{-3.4cm}
        \subfloat[First Choice]{\pgfplotsset{
    width = \textwidth,
    height = 0.628\textwidth
}

\begin{tikzpicture}[scale=1.0]

    \definecolor{expert0}{HTML}{F67F70}
    \definecolor{expert1}{HTML}{8ECFC9}
    \definecolor{expert2}{HTML}{FFBE7A}
    \definecolor{expert3}{HTML}{D2691E}
    \definecolor{expert4}{HTML}{7D9EEB}
    \definecolor{expert5}{HTML}{7FEA9E}
    \definecolor{expert6}{HTML}{FBD98D}
    \definecolor{expert7}{HTML}{20B2AA}

    \begin{axis}[
        ybar=0pt,
        xmin=0.5, xmax=3.5,
        ymin=0, ymax=100,
        xticklabels={SlimOrca, Magicoder, MetaMathQA},
        xtick={1,2,3},
        xtick align=inside,
        xticklabel style = {rotate=0}, 
        xticklabel style = {font=\tiny},
        ytick={0,20,40,60,80,100},
        yticklabel style = {font=\footnotesize},
        bar width = 4pt,
        ylabel near ticks,
        legend style={
            draw=none,
            at={(0.5,1.1)},
            anchor=south,
            legend columns=8,
            font=\scriptsize
        }
        ]
        \addplot+[ybar, fill=expert0, draw=black, area legend]         
        table [x={Expert},  y={Data}]     {data/First_Expert0.dat};
        \addplot+[fill=expert1, draw=black, area legend]         
        table [x={Expert},  y={Data}]     {data/First_Expert1.dat};
        \addplot+[fill=expert2, draw=black, area legend]         
        table [x={Expert},  y={Data}]     {data/First_Expert2.dat};
        \addplot+[fill=expert3, draw=black, area legend]         
        table [x={Expert},  y={Data}]     {data/First_Expert3.dat};
        \addplot+[fill=expert4, draw=black, area legend]         
        table [x={Expert},  y={Data}]     {data/First_Expert4.dat};
        \addplot+[fill=expert5, draw=black, area legend]         
        table [x={Expert},  y={Data}]     {data/First_Expert5.dat};
        \addplot+[fill=expert6, draw=black, area legend]         
        table [x={Expert},  y={Data}]     {data/First_Expert6.dat};
        \addplot+[fill=expert7, draw=black, area legend]   
        table [x={Expert},  y={Data}]     {data/First_Expert7.dat};
        \legend{Expert0, Expert1, Expert2, Expert3, Expert4, Expert5, Expert6, Expert7}
    \end{axis}
\end{tikzpicture}
        \label{fig:1st_choice}}
    \end{minipage}
    \begin{minipage}[]{0.318\linewidth}
        \vspace{0.445cm}
        \hspace{0.1cm}
        \subfloat[Second Choice]{\pgfplotsset{
    width = \textwidth,
    height = 0.628\textwidth
}

\begin{tikzpicture}[scale=1.0]
    \definecolor{expert0}{HTML}{F67F70}
    \definecolor{expert1}{HTML}{8ECFC9}
    \definecolor{expert2}{HTML}{FFBE7A}
    \definecolor{expert3}{HTML}{D2691E}
    \definecolor{expert4}{HTML}{7D9EEB}
    \definecolor{expert5}{HTML}{7FEA9E}
    \definecolor{expert6}{HTML}{FBD98D}
    \definecolor{expert7}{HTML}{20B2AA}
    
    \begin{axis}[
        ybar=0pt,
        xmin=0.5, xmax=3.5,
        ymin=0, ymax=50,
        xticklabels={SlimOrca, Magicoder, MetaMathQA},
        xtick={1,2,3},
        xtick align=inside,
        xticklabel style = {rotate=0}, 
        xticklabel style = {font=\tiny},
        ytick={0,10,20,30,40,50},
        yticklabel style = {font=\footnotesize},
        bar width = 4pt,
        ylabel near ticks,
        legend style={
            draw=none,
            at={(0.5,1.1)},
            anchor=south,
            legend columns=4,
        }
        ]
        \addplot+[ybar, fill=expert0, draw=black, area legend]         
        table [x={Expert},  y={Data}]     {data/Second_Expert0.dat};
        \addplot+[fill=expert1, draw=black, area legend]         
        table [x={Expert},  y={Data}]     {data/Second_Expert1.dat};
        \addplot+[fill=expert2, draw=black, area legend]         
        table [x={Expert},  y={Data}]     {data/Second_Expert2.dat};
        \addplot+[fill=expert3, draw=black, area legend]         
        table [x={Expert},  y={Data}]     {data/Second_Expert3.dat};
        \addplot+[fill=expert4, draw=black, area legend]         
        table [x={Expert},  y={Data}]     {data/Second_Expert4.dat};
        \addplot+[fill=expert5, draw=black, area legend]         
        table [x={Expert},  y={Data}]     {data/Second_Expert5.dat};
        \addplot+[fill=expert6, draw=black, area legend]         
        table [x={Expert},  y={Data}]     {data/Second_Expert6.dat};
        \addplot+[fill=expert7, draw=black, area legend]   
        table [x={Expert},  y={Data}]     {data/Second_Expert7.dat};
    \end{axis}
\end{tikzpicture}
        \label{fig:2nd_choice}}
    \end{minipage}
\caption{Proportion of tokens assigned to each expert on different dataset subsets.
}
\label{fig:router_analysis}
\end{figure*}

\begin{table}[t]
\centering
\resizebox{\linewidth}{!}{
\begin{tabular}{lc|c|cccccc}
\toprule
Model & \# Params & Avg. & Code & Math & \makecell[c]{CR} & \makecell[c]{WK} & MMLU\\ 
\midrule
Llama2-7B-Chat        & 7B & 35.4 & 14.9 & 15.1 & 66.7 & 33.0 & 47.3 \\ 
Vicuna-7B             & 7B & 34.0 & 9.6 & 13.5 & 67.6 & 29.2 & \textbf{50.1} \\
\midrule
Camelidae-8$\times$7B & 8B & \textbf{39.9} & \textbf{20.9} & \textbf{24.9} & \textbf{70.7} & 34.4 & 48.3 \\
\midrule[0.8pt]
Llama2-13B-Chat       & 13B & 41.8 & 23.1 & 21.2 & 70.9 & 40.0 & 53.8 \\
Vicuna-13B            & 13B & 39.9 & 10.7 & 21.0 & 70.8 & 41.1 & \textbf{55.8} \\
\midrule
Camelidae-8$\times$13B & 15B & \textbf{46.5} & \textbf{30.5} & \textbf{30.7} & \textbf{73.8} & \textbf{43.1} & 54.4 \\
\midrule[0.8pt]
Yi-34B-Chat                   & 34B & 51.8 & 30.4 & 42.5 & 73.3 & 38.0 & 74.8 \\
SUSChat-34B                   & 34B & 53.3 & 25.9 & 47.2 & 78.8 & 38.3 & \textbf{76.4} \\
\midrule
Camelidae-8$\times$34B & 38B & 59.3 & 42.7 & 50.5 & \textbf{79.7} & \textbf{47.8} & 75.6 \\
Camelidae-8$\times$34B-pro & 38B & \textbf{59.9} & \textbf{46.0} & \textbf{51.7} & 79.2 & 46.9 & 75.7 \\
\bottomrule
\end{tabular}}
\caption{Overall performance on grouped benchmarks of various dense models (Llama2-Chat \cite{touvron2023llama2}, Vicuna \cite{zheng2023judging}, Yi-Chat \cite{2023Yi}, SUSChat \cite{2023SUSChat}) across different model sizes.
We bold the highest scores separately for different model sizes.}
\label{table:main_3}
\end{table}

\minisection{Dense models vs. Sparse Models.}
We evaluate the efficacy of our novel training methodology through a comparative analysis of Camelidae models, encompassing both dense and sparse configurations across various parameter sizes, as delineated in \Cref{table:main_2} and \Cref{table:main_3}.
Camelidae models demonstrate a significant advantage over counterparts across different model sizes. 
This superiority is particularly evident in tasks requiring a deeper understanding, including code and mathematical benchmarks, highlighting the efficacy of our training approach in augmenting model capabilities.
To ensure equitable comparisons, Camel and Camelidae models were fine-tuned using the same dataset, IDAE-500K. 
As indicated in \Cref{table:main_2}, the Camelidae models, as sparse models, consistently display superior performance over the dense Camel models of comparable sizes. 
Moreover, Camelidae-8x34B-pro, which is trained utilizing the IDAE-720K dataset, outperforms Camelidae-8x34B which indicates that the effectiveness of our method is sustained even with the increment of the training data volume.

\minisection{Numbers of Experts.}
The results from the study, as shown in \Cref{table:ab_study_expert}, clearly demonstrate that increasing the number of experts in the MoE layers significantly enhances the model's performance. 
This trend is evident in the progressive improvement in scores across various academic benchmarks as the number of experts increases from 4 to 16 in the Camelidae models. 
Notably, the Camelidae-16$\times$7B model exhibits exceptional performance on all the benchmarks. 
This positive correlation between the number of experts and the model's performance indicates the untapped potential of our approach.
Specifically, a further increase in the number of experts might yield even more substantial advancements in model performance.

\subsection{Routing Analysis}

Our study rigorously examined the expert selection process by the router, with a keen focus on ascertaining whether specific experts demonstrate specialization in distinct domains such as coding and mathematics. 

This inquiry involved a thorough analysis of the distribution patterns of selected experts across various dataset subsets. 
These included SlimOrca \cite{SlimOrca, mukherjee2023orca, longpre2023flan}, Magicoder \cite{wei2023magicoder}, and MetaMathQA \cite{yu2023metamath}. 
The outcomes of this analysis are depicted in \Cref{fig:router_analysis}, with particular emphasis on the 15th layers of the Camelidae-8$\times$7B model.

Our findings highlight discernible variations in the distribution of experts among the three datasets.
For instance, Expert 1 exhibits a notably higher activation within the Magicoder dataset, while Expert 6 demonstrates a significant activation rate in the MetaMathQA dataset relative to other experts.
These observations suggest that the router operates with a structured syntactic approach. 
Importantly, despite the variation in expert selection across different datasets, certain experts (specifically Experts 1, 2, 5, and 6) consistently exhibit elevated activation rates.

\begin{table}[tb!]
\centering
\resizebox{0.92\linewidth}{!}{
\begin{tabular}{lc|c|cccccc}
\toprule
Model & \# Experts & Avg. & Code & Math & \makecell[c]{CR} & \makecell[c]{WK} & MMLU \\ 
\midrule
Camelidae-4$\times$7B & 4 & 39.6 & 20.7 & 24.3 & 70.2 & 33.3 & 49.3 \\
Camelidae-8$\times$7B & 8 & 39.9 & 20.9 & 24.9 & \textbf{70.7} & 34.4 & 48.3 \\
Camelidae-16$\times$7B & 16 & \textbf{40.5} & \textbf{21.6} & \textbf{25.8} & \textbf{70.7} & \textbf{35.0} & \textbf{49.4} \\
\bottomrule
\end{tabular}}
\caption{Evaluation on different numbers of experts in the MoE layers.
We bold the highest scores for each grouped benchmark.}
\label{table:ab_study_expert}
\end{table}



\section{Related Work}

\subsection{Dense and Sparse Models}
Traditional dense models activate all parameters during training and inference, leading to high computational and memory requirements as model sizes increase. 
In contrast, sparse models, employing the MoE architecture \cite{shazeer2017outrageously}, activate only a subset of the total available parameters for each input token.
In sparse models, the FFN layer is replaced by an MoE layer, directing each input token to a select group of expert networks for processing.
The final token representation is an amalgamation of outputs from these chosen experts.
Despite an increase in parameters, the sparse activation of experts ensures computational efficiency while enhancing model capabilities.
The sparse models with MoE architecture have been extensively explored in the field of NLP \cite{lepikhin2020gshard, du2022glam, fedus2022switch}, particularly with its integration into the transformer block.
Our approach adopts the routing strategy from \cite{lepikhin2020gshard, du2022glam}, with selective parameter activation to achieve computational efficiency.

\subsection{Reuse of Trained Weights}

Recent studies have focused on improving training efficiency by leveraging pre-existing model weights for a warm start, thus minimizing training expenses \cite{chen2015net2net, rae2021scaling, yang2021speeding, lin2021m6, lan2019albert}. 
Sparse Upcycling \cite{komatsuzaki2022sparse} introduces a methodology to initialize sparse MoE models using weights from a pre-trained dense model.
This approach significantly reduces the computational resources needed compared to the training of the original dense model. 
Sparse Upcycling involves the direct transfer of layer normalization, attention, and embedding parameters from the dense model to the new sparse model. 
Moreover, it replaces some Multilayer Perceptron (MLP) layers with MoE layers, initializing the experts in these layers with weights from the dense model's MLP.
This process effectively transfers valuable learned representations from the dense model's pre-training phase into the sparse model. 
In our research, we adopt this method, reusing weights from a pre-trained dense model for our PESC method.

\subsection{Parameter-Efficient Fine-Tuning}

Traditionally, full fine-tuning has been the norm for adapting pre-trained models, including LLMs. 
However, due to the immense size of LLMs, this approach demands substantial computational resources. 
To mitigate this, numerous PEFT methods have emerged \cite{houlsby2019parameter, hu2021lora, Li2021PrefixTuningOC, liu2022few, wu2024padapter}.
PEFT focuses on training a limited subset of parameters, either from the existing model or newly added ones. 
Adapter-based methods \cite{houlsby2019parameter, hu2021lora, liu2022few, wu2024padapter} integrate small, learnable modules called adapters into pre-trained models, fine-tuning only these newly inserted parameters. 
Among these, QLoRA \cite{dettmers2024qlora} has gained popularity for its efficiency in fine-tuning LLMs, yielding results comparable to full fine-tuning. 
 Another emerging trend in PEFT is prefix-/prompt-tuning \cite{lester-etal-2021-power, Li2021PrefixTuningOC}, involving the addition of learnable token vectors to either the keys and values in attention modules or directly to the input sequence.
In this study, we insert adapters after the copied FFN layers to construct MoE layers and employ QLoRA to update the other weight metrics of LLMs.

\subsection{Mixture of LoRA Experts}

Other works also explore the combination of MoE with PEFT techniques \cite{diao2023mixda, gou2023mclora, wu2024mole, liu2023moelora, luo2024moelora, dou2024loramoe}. 
For instance, LoRAMoE \cite{dou2024loramoe} focuses on the retention of world knowledge, and MoELoRA \cite{luo2024moelora} focuses on the Math and CommonSense Reasoning ability utilizing PEFT frameworks which unify MOE and LoRA.
However, the mixture of LoRA framework incurs additional computational costs including higher memory usage and slower speed without parallelism during the training and inference process. 
Our PESC method, in contrast, does not face these challenges.
PESC builds on the adapter-based model framework, fine-tuning multiple adapters inserted after the copied FFN layers instead of all the copied FFN layers in corresponding experts. 
In our MoE design of PESC, each expert utilizes a single adapter module, significantly reducing the overall memory footprint compared to LoRA module, which would require multiple modules per expert due to its placement in FFN and attention layers. 
This distinction is particularly crucial when dealing with a large number of experts, as memory constraints become increasingly challenging.
Moreover, our adapter-based experts enable parallel computation across experts due to their independence from each other's outputs, unlike LoRA, where dependencies between layers could limit parallelism. 
This design accelerates training time, especially in scenarios where the number of experts grows large, ensuring scalability and efficiency.
It's also worth noting that LoRA might require merging weights into the main model for inference, leading to increased memory usage and potential latency issues, especially since multiple tokens activate different experts. On the contrary, the adapter-based parameter-efficient MoE does not impose such overhead during inference, maintaining a low computational burden similar to the original dense model. 

\section{Conclusion}

In this paper, we introduce Parameter-Efficient Sparsity Crafting (PESC) which upcycles dense models into sparse models utilizing the MoE architecture.
PESC incorporates adapters \cite{houlsby2019parameter} within the MoE layers of sparse models, enabling the differentiation of experts without modifying the individual weights of each expert, and guarantees the quality of the approximation compared to traditional sparsity upcycling \cite{komatsuzaki2022sparse} in function space (\Cref{sec:pesc}). 
This technique significantly reduces computational costs and GPU memory requirements compared to sparse upcycling. 
It facilitates the expansion of model capacity with a minimal parameter increase due to the integration of adapters. 
We apply the PESC method to instruction tuning across various general tasks, resulting in notable performance enhancements on various benchmarks (\Cref{sec:exp}). 
Additionally, we develop sparse models, Camelidae, using the PESC approach and achieve superior performance across various open-source sparse models and demonstrate superior general capabilities compared to GPT-3.5.

\section*{Limitation}
The PESC method introduces slightly more parameters compared to some PEFT techniques (LoRA, etc.). 
The instruction tuning process of the sparse models utilizing the PESC method would require more GPU memory and computation time compared to dense models.
Although PESC enhances the performance of instruction tuning for general tasks, it may still not match the performance of sparse upcycling with full fine-tuning,
as PESC is a mathematical approximation of sparse upcycling as illustrated in \Cref{eq:err}.

\section*{Acknowledgement}

This work is partially supported by
The Research Grants Council of Hong Kong SAR (No.~CUHK14210723 and No.~CUHK14211824),
and the MIND project (MINDXZ202404).

\bibliography{ref/ref}

\clearpage
\appendix
\section{Details of IDAE Datasets}
\label{sec:appendix1}
We show the proportion of SlimORCA \cite{SlimOrca, mukherjee2023orca, longpre2023flan}, Magicoder \cite{wei2023magicoder}, and MetaMathQA \cite{yu2023metamath} datasets in IDAE-500K and IDAE-720K datasets in \Cref{table:datasets}.

\begin{table}[!htb]
\centering
\resizebox{0.828\linewidth}{!}{
\begin{tabular}{cccc}
\toprule
& SlimOrca & Magicoder & MetaMathQA \\ 
\midrule
IDAE-500K & 300K & 100K & 100K \\
IDAE-720K & 360K & 180K & 180K \\
\bottomrule
\end{tabular}}
\caption{The proportion of SlimORCA, Magicoder, and MetaMathQA datasets in IDAE datasets.}
\label{table:datasets}
\end{table}

\section{Implementation Details}
\label{sec:appendix2}
We show the hyperparameters that we use for instruction tuning in \Cref{table:hyperpara}.
\begin{table}[!htb]
\centering
\resizebox{\linewidth}{!}{
\begin{tabular}{cccccc}
\toprule
lr & epoch & LoRA\ $r$ & LoRA\ $\alpha$ & Quant Type & Adapter Dim \\
\midrule
$2 \times 10^{-4}$ & 1 & 64 & 16 & nf4 & 512\\
\bottomrule
\end{tabular}}
\caption{Hyperparameters of instruction tuning.}
\label{table:hyperpara}
\end{table}



\section{Detailed Evaluation Results on Grouped Benchmarks.}
\label{sec:appendix3}
We show the detailed evaluation results of each grouped academic benchmark as follows:
\begin{itemize}[itemsep=0pt,topsep=0pt,parsep=0pt]
    \item In \Cref{table:mmlu}, we report the evaluation details of the MMLU benchmark.
    \item In \Cref{table:math}, we report the results on GSM8K and MATH benchmarks.
    \item In \Cref{table:code}, we compare the results on HumanEval and MBPP benchmarks.
    \item In \Cref{table:cr}, we show the results on several commonsense reasoning benchmarks.
    \item In \Cref{table:wk}, We evaluate the performance on NaturalQuestions and TriviaQA benchmarks.
\end{itemize}

\begin{table}[!htbp]
\centering
\small
\resizebox{\linewidth}{!}{
\begin{tabular}{lccccc}
\toprule
& Humanities & STEM & Social Sciences & Other & Average \\
\midrule
LLaMA2-7B  & 43.2 & 36.9 & 51.7 & 52.6 & 45.7 \\
LLaMA2-7B-Chat  & 43.4 & 38.7 & 54.7 & 54.6 & 47.3 \\
Vicuna-7B  & 46.0 & 40.4 & 58.2 & 58.1 & 50.1 \\
Camel-7B  & 43.9 & 38.5 & 55.9 & 54.6 & 47.7 \\
Camelidae-8$\times$7B  & 44.7 & 38.1 & 56.9 & 55.9 & 48.3 \\
\midrule[0.8pt]
LLaMA2-13B & 52.3 & 44.1 & 63.7 & 62.0 & 55.1 \\
LLaMA2-13B-Chat & 50.3 & 43.9 & 62.6 & 60.3 & 53.8 \\
Vicuna-13B & 52.1 & 44.6 & 65.3 & 63.5 & 55.8 \\
Camel-13B & 52.0 & 42.2 & 63.0 & 61.7 & 54.4 \\
Camelidae-8$\times$13B & 52.1 & 43.3 & 62.6 & 61.1 & 54.4 \\
\midrule[0.8pt]
Yi-34B     & 71.3 & 67.3 & 85.4 & 80.2 & 75.5 \\
Yi-34B-Chat     & 70.5 & 66.3 & 84.7 & 79.9 & 74.8 \\
SUSChat-34B     & 72.2 & 69.6 & 85.5 & 80.5 & 76.4 \\
Camel-34B & 72.5 & 67.3 & 84.0 & 79.3 & 75.3 \\
Camelidae-8$\times$34B & 72.8 & 66.7 & 83.8 & 80.4 & 75.6 \\
Camelidae-8$\times$34B-pro & 73.8 & 66.0 & 83.8 & 80.3 & 75.7 \\
\bottomrule
\end{tabular}}
\caption{Comparison on the performance of MMLU.}
\label{table:mmlu}
\end{table}

\begin{table}[]
\centering
\small
\resizebox{0.648\linewidth}{!}{
\begin{tabular}{l|ccc}
\toprule
& GSM8K & MATH & Average\\
\midrule
LLaMA2-7B  & 16.7 & 3.3 & 10.0 \\
LLaMA2-7B-Chat  & 16.7 & 3.3 & 10.0 \\
Vicuna-7B  & 16.7 & 3.3 & 10.0 \\
Camel-7B   & 40.7 & 4.8 & 22.8 \\
Camelidae-8$\times$7B  & 44.0 & 5.8 & 24.9 \\
\midrule[0.8pt]
LLaMA2-13B & 29.6 & 5.0 & 17.3\\
LLaMA2-13B-Chat  & 16.7 & 3.3 & 10.0 \\
Vicuna-13B  & 16.7 & 3.3 & 10.0 \\
Camel-13B  & 50.2 & 8.4 & 29.3\\
Camelidae-8$\times$13B & 52.6 & 9.8 & 30.7 \\
\midrule[0.8pt]
Yi-34B     & 67.9 & 15.9 & 41.9\\
Yi-34B-Chat  & 16.7 & 3.3 & 10.0 \\
SUSChat-34B  & 16.7 & 3.3 & 10.0 \\
Camel-34B  & 76.1 & 18.2 & 47.2\\
Camelidae-8$\times$34B & 78.3 & 22.6 & 50.5\\
\bottomrule
\end{tabular}}
\caption{Comparison on mathematical reasoning tasks.}
\label{table:math}
\end{table}

\begin{table}[]
\centering
\small
\resizebox{0.648\linewidth}{!}{
\begin{tabular}{l|ccc}
\toprule
& HumanEval & MBPP & Average \\
\midrule
LLaMA2-7B  & 12.8 & 14.8 & 13.8 \\
LLaMA2-7B-Chat  & 16.7 & 3.3 & 10.0 \\
Vicuna-7B  & 16.7 & 3.3 & 10.0 \\
Camel-7B   & 17.7 & 21.0 & 19.4 \\
Camelidae-8$\times$7B  & 18.3 & 23.4 & 20.9 \\
\midrule[0.8pt]
LLaMA2-13B & 18.9 & 26.8 & 22.9 \\
LLaMA2-13B-Chat  & 16.7 & 3.3 & 10.0 \\
Vicuna-13B  & 16.7 & 3.3 & 10.0 \\
Camel-13B  & 28.7 & 30.3 & 29.5 \\
Camelidae-8$\times$13B & 30.6 & 30.4 & 30.5 \\
\midrule[0.8pt]
Yi-34B     & 26.2 & 38.2 & 32.2 \\
Yi-34B-Chat  & 16.7 & 3.3 & 10.0 \\
SUSChat-34B  & 16.7 & 3.3 & 10.0 \\
Camel-34B  & 42.1 & 40.6 & 41.4 \\
Camelidae-8$\times$34B & 43.9 & 41.4 & 42.7 \\
\bottomrule
\end{tabular}}
\caption{Comparison on code generation tasks.}
\label{table:code}
\end{table}

\begin{table}[tb!]
\centering
\resizebox{\linewidth}{!}{
\begin{tabular}{l|cccccc}
\toprule
& PIQA & HellaSwag & WinoGrande & ARC-e & ARC-c & Average \\
\midrule
LLaMA2-7B  & 78.9 & 75.9 & 69.5 & 74.7 & 46.2 & 69.0 \\
LLaMA2-7B-Chat  & 77.0 & 75.5 & 66.4 & 69.7 & 44.7 & 66.7 \\
Vicuna-7B   & 78.0 & 73.7 & 69.3 & 71.3 & 45.8 & 67.6 \\
Camel-7B   & 79.7 & 76.8 & 71.3 & 75.0 & 47.9 & 70.1 \\
Camelidae-8$\times$7B  & 79.9 & 76.8 & 72.1 & 75.0 & 49.6 & 70.7 \\
\midrule[0.8pt]
LLaMA2-13B & 80.7 & 80.8 & 71.9 & 77.4 & 48.9 & 71.6 \\
LLaMA2-13B-Chat & 79.1 & 79.7 & 71.3 & 73.8 & 50.3 & 70.9 \\
Vicuna-13B  & 78.9 & 77.4 & 71.9 & 74.8 & 50.9 & 70.8 \\
Camel-13B  & 80.9 & 79.8 & 74.6 & 77.7 & 54.3 & 73.5 \\
Camelidae-8$\times$13B & 80.9 & 80.1 & 74.7 & 78.8 & 54.2 & 73.8 \\
\midrule[0.8pt]
Yi-34B     & 82.9 & 83.7 & 78.9 & 84.1 & 61.6 & 78.2 \\
Yi-34B-Chat     & 79.9 & 80.7 & 77.1 & 74.3 & 54.6 & 73.3 \\
SUSChat-34B  & 82.0 & 83.0 & 81.0 & 84.8 & 63.0 & 78.8 \\
Camel-34B  & 82.3 & 82.6 & 80.0 & 86.1 & 63.6 & 78.9 \\
Camelidae-8$\times$34B & 82.7 & 83.2 & 80.9 & 86.2 & 65.2 & 79.7 \\
Camelidae-8$\times$34B-pro & 83.6 & 82.5 & 80.1 & 86.6 & 63.3 & 79.2 \\
\bottomrule
\end{tabular}}
\caption{Comparison on the performance of various commonsense reasoning tasks.}
\label{table:cr}
\end{table}

\begin{table}[tb!]
\centering
\resizebox{0.688\linewidth}{!}{
\begin{tabular}{l|ccc}
\toprule
& NaturalQuestions & TriviaQA & Average \\
\midrule
LLaMA2-7B  & 19.1 & 52.8 & 36.0 \\
LLaMA2-7B-Chat  & 19.6 & 46.4 & 33.0 \\
Vicuna-7B   &  15.6 & 42.8 & 29.2 \\
Camel-7B   &  17.6 & 51.0 & 34.3 \\
Camelidae-8$\times$7B  & 17.8 & 51.0 & 34.4 \\
\midrule[0.8pt]
LLaMA2-13B & 24.8 & 59.4 & 42.1 \\
LLaMA2-13B-Chat & 25.0 & 55.0 & 40.0 \\
Vicuna-13B  & 25.8 & 56.3 & 41.1 \\
Camel-13B  & 24.7 & 57.5 & 41.1 \\
Camelidae-8$\times$13B & 26.8 & 59.4 & 43.1 \\
\midrule[0.8pt]
Yi-34B     & 33.5 & 62.1 & 47.8 \\
Yi-34B-Chat  & 23.7 & 52.3 & 38.0 \\
SUSChat-34B  & 20.4 & 56.1 & 38.3 \\
Camel-34B  & 31.6 & 63.3 & 47.5 \\
Camelidae-8$\times$34B & 32.2 & 63.4 & 47.8 \\
Camelidae-8$\times$34B-pro & 31.2 & 62.5 & 46.9 \\
\bottomrule
\end{tabular}}
\caption{Comparison on the exact match performance of world knowledge tasks.}
\label{table:wk}
\end{table}

\end{document}